\documentclass[runningheads]{llncs}

\usepackage[T1]{fontenc}
\usepackage{graphicx,verbatim}
\usepackage{url}
\usepackage{bbding}
\usepackage{amsmath,amsfonts}
\usepackage{xcolor}
\definecolor{darkgreen}{RGB}{0,140,90}  
\begin{document}
\title{Do Not Break the Vessels: Structure-Preserving Mean Flow for Vascular Image Translation}
\titlerunning{SPMF for Vascular Image Translation}
\begin{comment}  

\end{comment}

\author{
Changjin Sun\inst{1} \and
Zhuo Hu\inst{2} \and
Kaini Wang\inst{3} \and
Baixuan Wu\inst{4} \and
Shuo Gao\inst{1} \and
Runan Zheng\inst{5} \and
Cheng Xue\inst{2} \and
Yudong Zhang\textsuperscript{2(\Envelope)} \and
Guangquan Zhou\textsuperscript{1(\Envelope)}
}

\authorrunning{C. Sun et al.}

\institute{
School of Biological Science and Medical Engineering, Southeast University, Nanjing, China\\
\email{guangquan.zhou@seu.edu.cn}
\and
School of Computer Science and Engineering, Southeast University, Nanjing, China
\email{yudongzhang@ieee.org}
\and
Department of Computer Science and Engineering, The Chinese University of Hong Kong, Hong Kong, China
\and
Zhejiang University, Hangzhou, China
\and
Suzhou Microclear Medical Instruments Co., Suzhou, China
}

\maketitle              

\begin{abstract}
Reconstructing anatomically faithful vascular structures from clinically accessible imaging modalities is of substantial clinical significance. However, existing cross-modal translation methods mainly emphasize pixel-level fidelity or visual realism and treat structure preservation as a property of the final output rather than an invariant of the generative process. This limitation often leads to structural discontinuities and artifacts, compromising anatomical coherence and clinical reliability. In this work, we propose a \emph{Structure-Preserving Mean Flow (SPMF)} framework that formulates vascular image translation as a topology-invariant transport process. From a structural invariance principle, we derive an orthogonality constraint on the flow velocity field that formally separates appearance transport from topological distortion, and implement it as a time-weighted surrogate objective within a Brownian-bridge diffusion model to preserve topology at every diffusion step. Moreover, we propose a \emph{Prototype-Guided Structural Refinement (PGSR)} module to align degraded inference-time structures with reliable training-time structures. Experiments on paired NIRII–2PF and fundus datasets demonstrate consistent improvements over state-of-the-art methods, achieving peak PSNRs of 24.96~dB and 24.83~dB, respectively.


\end{abstract}

\section{Introduction}
\begin{figure}
\centering
\includegraphics[width=0.77\textwidth]{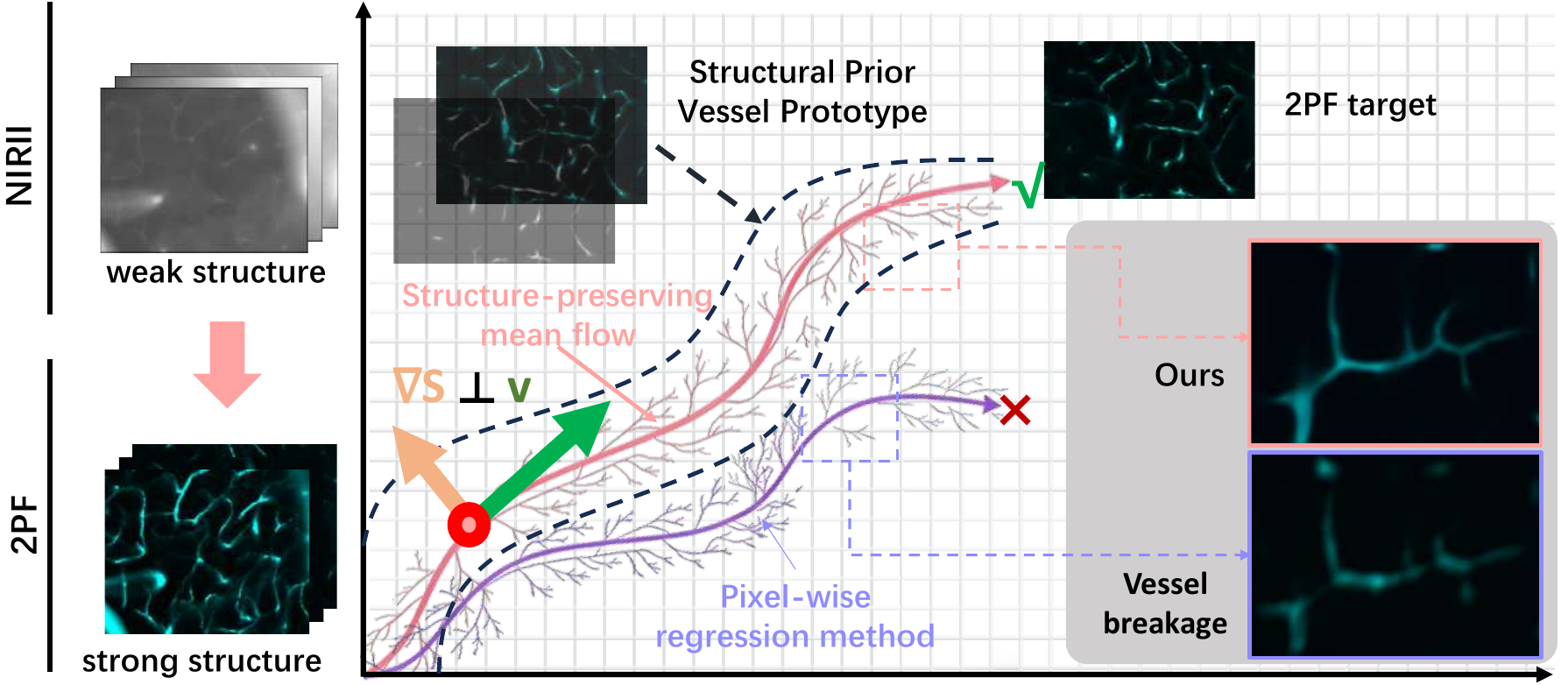}
\caption{Illustration of the proposed structure-preserving mean flow. By constraining the velocity field to be orthogonal to the structural gradient, the generative dynamics can transform appearance while remaining on the vascular structural manifold, thereby preserving topology along the entire trajectory.} \label{fig1}
\end{figure}

Microvascular topology is a key biomarker for diagnosing and monitoring vascular diseases~\cite{Cheung2015RetinalMicrovasculature}, making structure-faithful imaging a clinical necessity. Two-photon fluorescence (2PF) microscopy meets this requirement by resolving sub-micron vascular detail through nonlinear optical excitation, but its reliance on specialized pulsed lasers and point-scanning acquisition makes it impractical for routine clinical use. Near-infrared window II (NIRII) imaging offers a non-invasive, fast, and clinically compatible alternative~\cite{1,2,3,4,5,6}, yet severe tissue scattering and limited spatial resolution fundamentally degrade vascular topology in the acquired images (Fig.~1). Bridging the NIRII-to-2PF modality gap is therefore a structure-critical translation task that demands topologically faithful vascular recovery, not just visually plausible outputs.

Existing methods for cross-modal medical image translation, whether based on pixel-wise regression~\cite{7,8,9,10,11} or conditional generation~\cite{12,13,14}, treat structure preservation as a property of the final output rather than an invariant of the generative process itself. Pixel-wise methods focus on intensity fidelity, while diffusion-based approaches prioritize visual realism and leave the generative trajectory unconstrained, making topological violations such as vessel breakage and spurious branches admissible outcomes. This problem is further compounded by a distributional gap between training and inference. Structural cues extracted from degraded NIRII images differ substantially from the clean structures available during training, making any output-level structural supervision unreliable even when it is explicitly applied.

Vascular topology is an intrinsic anatomical property that remains invariant across imaging modalities despite substantial appearance variation~\cite{21,22,wang2022awsnet,wang2023dlgnet,wang2024sbcnet}. We argue that a valid translation must preserve this invariance not only at the output but throughout the generative process. When image translation is modeled as a continuous flow driven by a time-dependent velocity field, the vesselness of the evolving image should remain constant along the trajectory, which in turn implies that the velocity field must be orthogonal to the structural gradient. Standard mean flow formulations impose no such constraint, leaving the velocity field free to alter appearance and topology alike. This distinction motivates our structure-preserving mean flow, in which the generative dynamics may freely transform appearance but are prevented from crossing the boundary of the structural manifold (Fig.~1).

\begin{figure}[t]
\centering
\includegraphics[width=\textwidth]{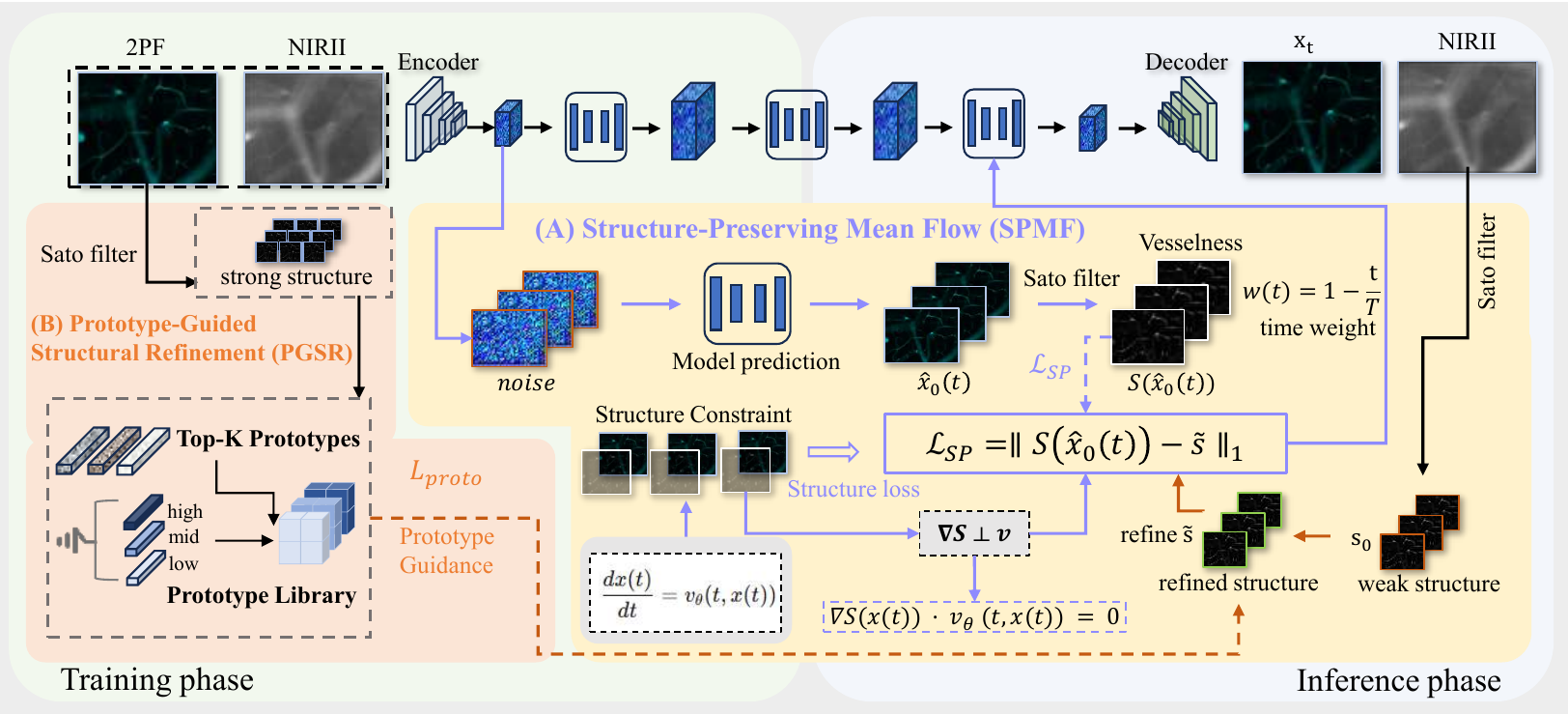}
\caption{Overview of the proposed framework. (A) SPMF guides diffusion-based translation for topology-consistent generation, while (B) PGSR aligns degraded input vesselness with reliable training-time anatomical patterns.}
\label{fig2}
\end{figure}

Following this formulation, we propose a \emph{Structure-Preserving Mean Flow (SPMF)} and instantiate it within a Brownian-bridge diffusion model~\cite{15}. At each diffusion step, the model predicts a clean reconstruction from the current noisy state, and we enforce that the predicted image maintains the same vascular topology as a reference anatomical prior throughout the trajectory. A time-dependent weighting schedule modulates this constraint according to the generative hierarchy of diffusion, imposing strong topology regularization during the early coarse-structure stage and progressively relaxing it as generation shifts to fine-grained appearance refinement. Since this per-step supervision requires a reliable structural reference, and vesselness maps extracted from degraded NIRII inputs are noisy and topologically incomplete, we further propose a \emph{Prototype-Guided Structural Refinement (PGSR)} mechanism that retrieves nearest-neighbor prototypes from a library of training-time vascular patterns and aligns the input vesselness map through frequency-decomposed optimization with topology-sensitive weighting, projecting both training and inference structures onto a shared anatomical manifold.

Our contributions are three-fold. (1) We derive an orthogonality constraint on the flow velocity field from a structural invariance principle, formally separating appearance transport from topological distortion, and implement it as a time-weighted surrogate objective that preserves topology at every diffusion step. (2) We propose a prototype-guided structural refinement that bridges the training–inference structural gap by projecting noisy vesselness estimates onto a shared anatomical manifold learned from training data. (3) Experiments on paired NIRII–2PF and fundus datasets demonstrate consistent improvements over state-of-the-art methods in both structural fidelity and visual quality, with peak PSNRs of 24.96~dB and 24.83~dB, respectively.

\section{Methodology}

The proposed \emph{Structure-Preserving Mean Flow (SPMF)} framework addresses NIRII-to-2PF translation from a topology-consistent flow perspective (Fig.~2). Section~2.1 formulates the problem setting and highlights the structural discrepancy between training and inference. Section~2.2 introduces the proposed SPMF, which constrains the generative dynamics to follow topology-consistent trajectories. Section~2.3 presents the \emph{Prototype-Guided Structural Refinement (PGSR)} module that constructs a unified anatomical representation to support structure-preserving generation.

\subsection{Problem Formulation}

Given paired data $\mathcal{D} = \{(y, x)\}$, where $y$ denotes a NIRII image and $x$ the corresponding 2PF image, our goal is to learn a conditional model $p_\theta(x \mid y)$ that recovers 2PF-like appearance while preserving vascular topology. We extract vesselness maps using the Sato operator $\mathcal{S}(\cdot)$~\cite{16,25}. During training, reliable structures $s^{\mathrm{2pf}} = \mathcal{S}(x)$ are available, while at inference only degraded estimates $s^{\mathrm{nir2}} = \mathcal{S}(y)$ can be obtained.

Due to severe noise and loss of high-frequency details in NIRII imaging, the distributions of $s^{\mathrm{2pf}}$ and $s^{\mathrm{nir2}}$ differ substantially. Directly enforcing a constraint such as $\mathcal{S}(\hat{x}) \approx \mathcal{S}(y)$ is therefore ill-posed and often leads to unstable optimization and structural artifacts. Our objective is thus to learn a conditional generative model that (i) preserves vascular topology throughout the generative process and (ii) bridges the discrepancy between reliable training-time structures and degraded inference-time estimates via a unified structural representation.

\subsection{Structure-Preserving Mean Flow (SPMF)}

The proposed SPMF formulates the generative process as a continuous-time transport constrained by anatomical invariance. It models image translation as a flow driven by a time-dependent velocity field while explicitly restricting the dynamics to follow topology-consistent trajectories. Concretely, the generative process is described as a continuous-time flow governed by a time-dependent velocity field and a structure-preserving constraint,
\begin{equation}
\frac{d x(t)}{dt} = v_\theta(t, x(t)).
\end{equation}

\subsubsection{Structure-Preserving Principle.}

Let $x(t) \in \mathbb{R}^{H \times W}$ denote the generated image at time $t$. We impose the following principle:

\noindent\textbf{Principle 1 (Structure Preservation).}
Along a valid generative trajectory, the underlying anatomical structure remains invariant:
\begin{equation}
\mathcal{S}(x(t)) = \mathcal{S}(x(0)), \quad \forall t \in [0,1].
\label{eq:structure_invariance}
\end{equation}
This reflects the nature of NIRII-to-2PF translation, where appearance changes substantially while vascular topology should remain unchanged.

\subsubsection{Constraint on Flow Dynamics.}

Taking the time derivative of Eq.~\eqref{eq:structure_invariance} yields
\begin{equation}
\frac{d}{dt} \mathcal{S}(x(t)) 
= \nabla_x \mathcal{S}(x(t)) \cdot v_\theta(t, x(t)).
\end{equation}
To preserve structure, the above term must vanish, leading to the constraint
\begin{equation}
\nabla_x \mathcal{S}(x(t)) \cdot v_\theta(t, x(t)) = 0,
\label{eq:orth_constraint}
\end{equation}
which implies that the velocity field must lie in the tangent space of the structure manifold, allowing appearance changes while preventing topological distortion.

\subsubsection{Surrogate Objective in Diffusion.}

Directly enforcing Eq.~\eqref{eq:orth_constraint} is intractable in practice due to the non-differentiability of the vesselness operator. We therefore adopt a surrogate objective based on structural consistency within a Brownian-bridge diffusion framework~\cite{15}.

Given a noisy sample $x_t$, the model predicts a reconstruction $\hat{x}_0(t)$. We enforce structure preservation by minimizing a time-weighted consistency loss:
\begin{equation}
\mathcal{L}_{\mathrm{SP}}(\theta)
=
\mathbb{E}_{t}
\left[
w(t)
\left\|
\mathcal{S}(\hat{x}_0(t)) - \tilde{s}
\right\|_1
\right],
\quad
w(t) = 1 - \frac{t}{T}.
\end{equation}
This weighting emphasizes structural consistency during early stages, when global topology is formed, and gradually relaxes the constraint to allow fine-grained appearance refinement. In this way, the diffusion process is guided to follow a structure-preserving mean flow.

\subsection{Prototype-Guided Structural Refinement (PGSR)}

PGSR provides a stable anatomical reference $\tilde{s}$ by projecting raw vesselness estimates onto a shared anatomical manifold learned from training data~\cite{22,27}. Given an initial vesselness map $s = \mathcal{S}(\cdot)$, a refinement operator is defined as
\begin{equation}
\tilde{s} = \mathcal{R}_{\mathrm{proto}}(s),
\end{equation}
which produces a structurally consistent representation shared across training and inference stages.

\subsubsection{Prototype Library and Retrieval.}

A prototype set $\mathcal{P} = {p_k}_{k=1}^K$ is constructed from vesselness maps of training samples, representing typical vascular patterns in terms of topology, thickness, and connectivity. For a given input structure $s$, the most similar prototypes are retrieved as references using a similarity measure in the vesselness space. These prototypes serve as anatomical anchors that regularize the refinement toward valid vascular configurations.

\subsubsection{Frequency-Decomposed Structural Alignment.}

Both the input structure $s$ and the retrieved prototype $p$ are decomposed into low-, mid-, and high-frequency components, denoted as $\{s^{(f)}, p^{(f)}\}$ with $f \in \{\text{low}, \text{mid}, \text{high}\}$. While NIRII/2PF observations exhibit large variations in low- and mid-frequency appearance, vascular topology is primarily encoded in high-frequency responses, such as thin branches and sharp boundaries. We therefore impose frequency-dependent weights and emphasize high-frequency alignment.

The refinement is obtained by minimizing the following objective:
\begin{equation}
\mathcal{L}_{\text{proto}} =
\sum_{f \in \{\text{low},\text{mid},\text{high}\}}
\alpha_f \left\| s^{(f)} - p^{(f)} \right\|_2^2
+ \lambda_{\text{tv}} \|\nabla s\|_1,
\end{equation}
where $(\alpha_{\text{low}}, \alpha_{\text{mid}}, \alpha_{\text{high}}) = (0.5, 1.0, 2.0)$, and the total variation term suppresses noise and encourages spatial continuity.

\subsubsection{Unified Refinement for Training and Inference.}

The above optimization is performed for a small number of iterations, yielding a refined structure $\tilde{s}$ that lies on a valid anatomical manifold. Importantly, the same refinement process is applied during both training and inference:
\begin{equation}
\tilde{s}^{\text{train}} = \mathcal{R}_{\mathrm{proto}}(\mathcal{S}(x)), \quad
\tilde{s}^{\text{test}} = \mathcal{R}_{\mathrm{proto}}(\mathcal{S}(y)).
\end{equation}
This unified formulation reduces the structural distribution gap between training and testing stages and provides a stable anatomical prior for the proposed structure-preserving mean flow.

\section{Experiments}
\subsubsection{Dataset and Experimental Setup.}

We evaluate the proposed method on a paired NIRII--2PF dataset (NIR2PF) with 763 image pairs and an external fundus dataset (Fundus) with 1001 patients. For Fundus, paired infrared (IR) and RGB images are used, where the IR image serves as a strong structural reference and the $4\times$ downsampled green channel is treated as a degraded structural observation. The Fundus experiment is used as an external generalization test under degraded structural observations, rather than as a direct replacement for the primary NIRII-to-2PF translation task. All images are normalized and split into training, validation, and test sets with a ratio of 6:2:2. We report PSNR and SSIM, and additionally use the Vesselness Mean Squared Error (V-MSE), defined as the mean squared error between the Sato vesselness responses of the generated and reference images, which serves as a practical proxy for structural accuracy by penalizing missing, blurred, or spurious vessel responses.

\subsubsection{Implementation Details.}

Our method is implemented based on a Brownian Bridge Diffusion Model (BBDM)\cite{15}.
All images are resized to $256 \times 256$ and normalized to $[0,1]$.
The model is trained for 200 epochs (300K steps) using the Adam optimizer with a learning rate of $1\times10^{-4}$.
Structure preservation is enforced via the proposed SPMF and prototype-guided refinement.

\subsubsection{Comparison with State-of-the-Art Methods.}

We compare the proposed method with several representative image-to-image translation and super-resolution approaches, including SRGAN\cite{17}, SRN\cite{11}\cite{19}, the Brownian Bridge Diffusion Model (BBDM)\cite{15}, LDL\cite{18}, and SelfRDB\cite{20}.
All competing methods are retrained on both datasets with official implementations and identical splits.

\begin{table*}[t]
\centering
\caption{Quantitative comparison on the NIR2PF and Fundus datasets. V-MSE is reported in units of $10^{-3}$.}
\label{tab:comparison}
\setlength{\tabcolsep}{8pt}
\small
\begin{tabular}{lccc|ccc}
\hline
 & \multicolumn{3}{c|}{\textbf{NIR2PF}} & \multicolumn{3}{c}{\textbf{Fundus}} \\
Method 
& PSNR$\uparrow$ & SSIM$\uparrow$ & V-MSE$\downarrow$ 
& PSNR$\uparrow$ & SSIM$\uparrow$ & V-MSE$\downarrow$ \\
\hline
SRGAN~\cite{17}    & 24.39 & 0.5691 & 5.657 & 24.23 & 0.4657 & 13.19 \\
SRN~\cite{11,19}   & \underline{24.49} & 0.6680 & 5.576 & 18.04 & 0.4570 & 13.63 \\
BBDM~\cite{15}     & 22.39 & 0.6382 & \underline{3.568} & \underline{24.47} & 0.5204 & 12.02 \\
LDL~\cite{18}     & 23.81 & \textbf{\textcolor{darkgreen}{0.7215}} & 3.634 & 24.32 & \underline{0.5412} & \underline{8.65} \\
SelfRDB~\cite{20}  & 20.88 & 0.5875 & 5.565 & 22.97 & 0.5325 & 10.13 \\
\textbf{Proposed} & \textbf{\textcolor{darkgreen}{24.96}} & \underline{0.6904} & \textbf{\textcolor{darkgreen}{2.499}} & \textbf{\textcolor{darkgreen}{24.83}} & \textbf{\textcolor{darkgreen}{0.5823}} & \textbf{\textcolor{darkgreen}{7.76}} \\
\hline
\end{tabular}
\end{table*}

\begin{figure}[t]
\centering
\includegraphics[width=0.9\textwidth]{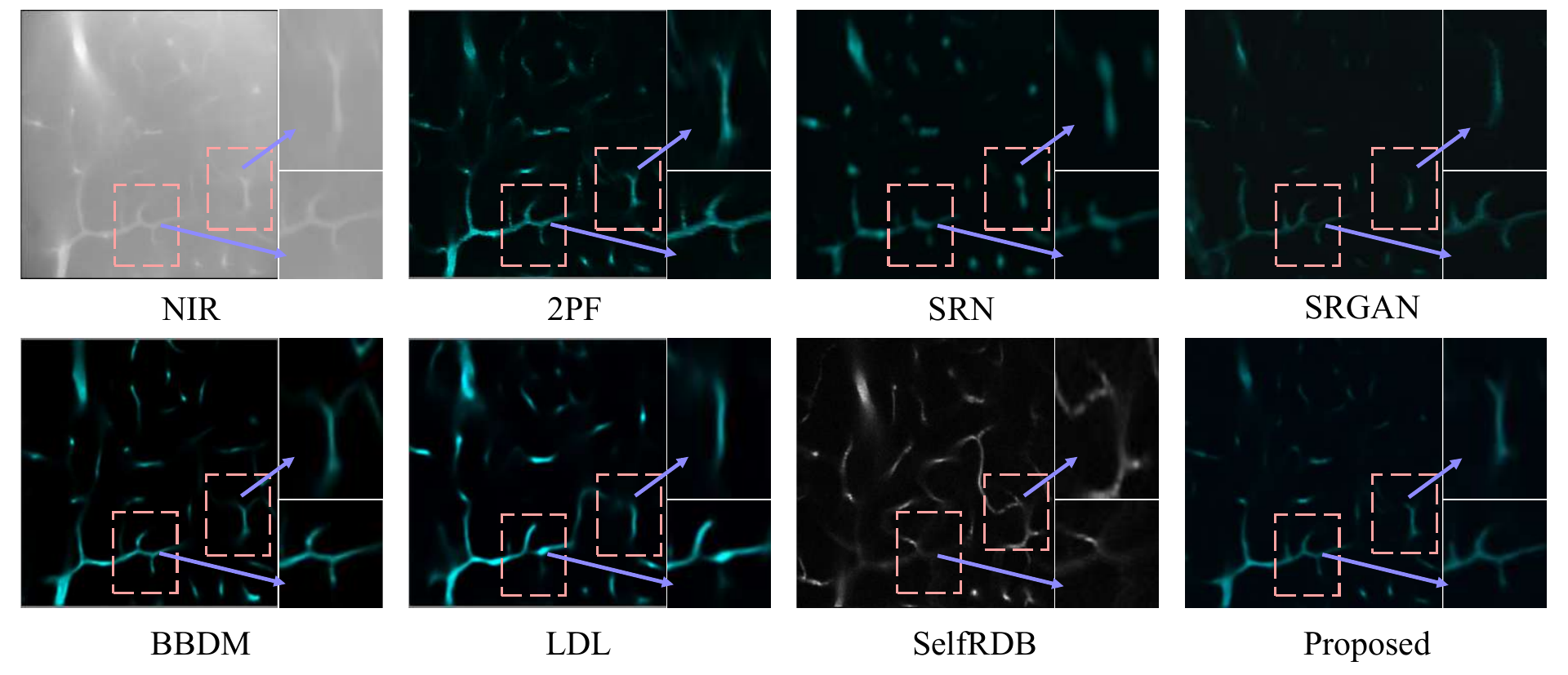}
\caption{Visual comparison of vascular image translation results on the NIR2PF dataset. Existing methods show vessel breakage, over-smoothing, or spurious structures, whereas the proposed method preserves topologically consistent vascular reconstructions with clearer continuity.}
\label{fig:visual}
\end{figure}

\begin{table}[t]
\centering
\setlength{\tabcolsep}{10pt} 
\caption{Ablation study of different components.}
\label{tab:ablation}
\begin{tabular}{c c c c c c}
\hline
BBDM & PGSR & SPMF & PSNR (dB)$\uparrow$ & SSIM$\uparrow$ & V-MSE($\times 10^{-3}$)$\downarrow$\\
\hline
$\checkmark$ &  &  & 22.39 & 0.6382 & 3.568 \\
$\checkmark$ & $\checkmark$ &  & 24.59 & 0.6769 & 3.124 \\
$\checkmark$ &  & $\checkmark$ & 23.31 & 0.6534 & 2.539 \\
$\checkmark$ & $\checkmark$ & $\checkmark$ & \textbf{\textcolor{darkgreen}{24.96}} & \textbf{\textcolor{darkgreen}{0.6904}} & \textbf{\textcolor{darkgreen}{2.499}}\\
\hline
\end{tabular}
\end{table}

\begin{figure}[t]
\centering
\includegraphics[width=\textwidth]{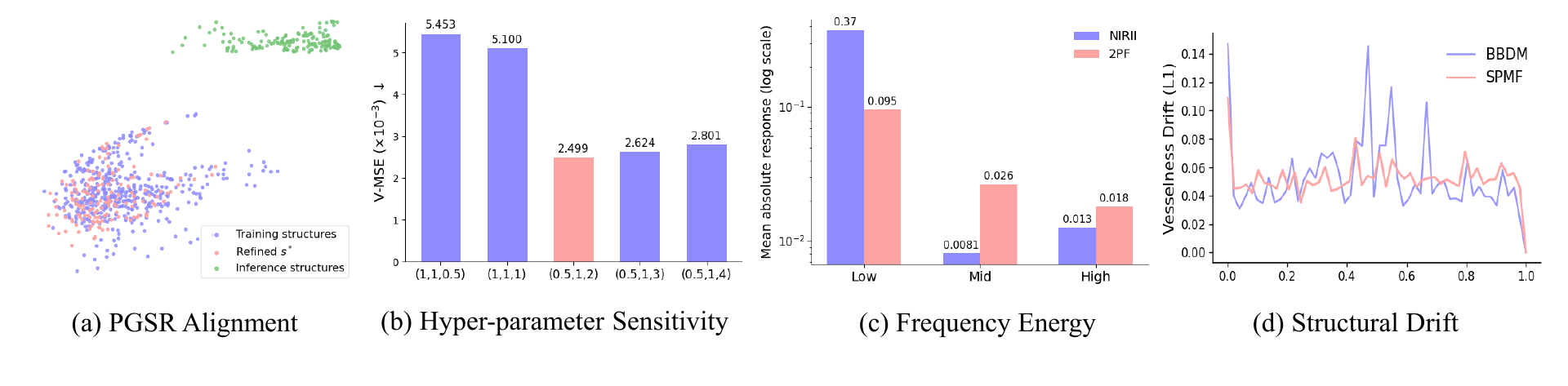}
\caption{Component-wise analysis of the proposed framework.
(a) PGSR and the training–inference structural gap. (b) Sensitivity to frequency weights. (c) Frequency-wise energy distribution. (d) Structural drift along the generative trajectory.}
\label{fig:analysis}
\end{figure}

The proposed method achieves the best or highly competitive performance on both the NIR2PF and Fundus datasets (Table~\ref{tab:comparison}). On NIR2PF, it attains the highest PSNR of 24.96~dB and the lowest V-MSE of $2.499\times 10^{-3}$, clearly surpassing the BBDM baseline, which achieves 22.39~dB PSNR and $3.568\times 10^{-3}$ V-MSE. Although LDL reports a slightly higher SSIM of 0.7215, this improvement is accompanied by noticeably worse structural accuracy, as reflected by its higher V-MSE of $3.634\times 10^{-3}$. On the Fundus dataset, the proposed method achieves the best PSNR of 24.83~dB, the best SSIM of 0.5823, and the lowest V-MSE of $7.76\times 10^{-3}$, demonstrating strong robustness under severely degraded structural observations.

\subsubsection{Visual Comparison.}

The proposed method consistently produces more topologically consistent and visually faithful vascular reconstructions (Fig.~3). Regression-based methods tend to over-smooth thin vessels and suppress fine branches, while diffusion-based methods without explicit structural constraints often introduce distorted or hallucinated patterns. Although LDL achieves a relatively high SSIM, its results exhibit evident over-smoothing and loss of fine branches. In contrast, the proposed method better preserves vessel continuity and bifurcation geometry, yielding results closer to the ground truth.

\subsubsection{Ablation Study.}

The ablation results on NIR2PF show that both PGSR and SPMF contribute to the overall performance (Table~\ref{tab:ablation}). Introducing PGSR yields a substantial improvement, raising PSNR from 22.39~dB to 24.59~dB and reducing V-MSE, which confirms the importance of a reliable anatomical prior aligned to the training manifold. Using SPMF alone also improves structural accuracy by constraining the generative dynamics, indicating the benefit of trajectory-level regularization even without explicit refinement. When both components are combined, the model achieves the best overall performance with 24.96~dB PSNR, 0.6904 SSIM, and the lowest V-MSE, clearly demonstrating their complementary effects.

\subsubsection{Component-wise Analysis.}

The following analysis examines the roles of PGSR and SPMF from both structural and dynamical perspectives (Fig.~4). Fig.~4(a) shows that refined structures cluster near the training manifold, whereas raw inference-time structures remain scattered, indicating that PGSR reduces the structural distribution gap. Fig.~4(b) evaluates the sensitivity to frequency weights and shows that balanced or low-frequency-dominated settings lead to higher errors, while emphasizing high-frequency components reduces the V-MSE to approximately $2.499 \times 10^{-3}$, supporting the frequency-weighted refinement design. Fig.~4(c) reports the frequency-wise vesselness energy distribution, where NIRII is dominated by low-frequency responses while 2PF retains non-negligible mid- and high-frequency components, confirming that fine vascular structures are mainly encoded at higher frequencies. Fig.~4(d) analyzes the structural drift along the generative trajectory and shows that SPMF maintains a more stable drift than the baseline, indicating that structural consistency is enforced throughout the diffusion process rather than only at the final output.

\section{Conclusion}

NIRII-to-2PF image translation is important for clinically accessible vascular visualization. We propose a structure-preserving mean flow framework that encourages topology-consistent diffusion dynamics by combining prototype-guided refinement with trajectory-level structural constraints. While the current evaluation mainly relies on vesselness-based structural errors, future work will incorporate explicit skeleton- or centerline-level metrics, larger datasets, and expert assessment to further validate clinical utility.

\vspace{1em}
\subsubsection{Acknowledgments.}
This work was supported in part by the National Key Research and Development Program of China (2022YFC2404300, 2024YFF1206700).

\vspace{1em}
\subsubsection{Disclosure of Interests.}
The authors have no competing interests to declare that are relevant to the content of this article.

\bibliographystyle{splncs04}
\bibliography{References}

\end{document}